\documentclass[conference]{IEEEtran}
\usepackage{blindtext, graphicx}
\usepackage{balance}

\ifCLASSINFOpdf
\else
\fi

\begin{document}

%
\title{Deep Learning applied to NLP}

\author{\IEEEauthorblockN{Marc Moreno Lopez}
\IEEEauthorblockA{College of Engineering and\\Applied Sciences\\
University of Colorado Colorado Springs\\
Colorado Springs, Colorado \\
Email: mmorenol@uccs.edu}
\and
\IEEEauthorblockN{Jugal Kalita}
\IEEEauthorblockA{College of Engineering and\\Applied Sciences\\
University of Colorado Colorado Springs\\
Colorado Springs, Colorado \\
Email: jkalita@uccs.edu}}

\maketitle

\begin{abstract}
Convolutional Neural Network (CNNs) are typically associated with Computer Vision. CNNs are responsible for major breakthroughs in Image Classification and are the core of most Computer Vision systems today. More recently CNNs have been applied to problems in Natural Language Processing and gotten some interesting results. In this paper, we will try to explain the basics of CNNs, its different variations and how they have been applied to NLP.
\end{abstract}


%
\IEEEpeerreviewmaketitle

\section{Introduction}
Deep learning methods are becoming important due to their demonstrated success at tackling complex learning problems. At the same time, increasing access to high-performance computing resources and state-of-the-art open-source libraries are making it more and more feasible for everyone to use these methods.

Natural Language Processing focuses on the interactions between human language and computers. It sits at the intersection of computer science, artificial intelligence, and computational linguistics. 
NLP is a way for computers to analyze, understand, and derive meaning from human language in a smart and useful way. By utilizing NLP, developers can organize and structure knowledge to perform tasks such as automatic summarization, translation, named entity recognition, relationship extraction, sentiment analysis, speech recognition, and topic segmentation.
The development of NLP applications is challenging because computers traditionally require humans to communicate to them via a programming language. Programming languages are precise, unambiguous and highly structured. Human speech, however, is not always precise, it is often ambiguous and the linguistic structure can depend on many complex variables, including slang, regional dialects and social context.

\subsection{Introduction to CNN}
A Neural Network is a biologically-inspired programming paradigm which enables a computer to learn from observed data. It is composed of a large number of interconnected processing elements, neurons, working in unison to solve a problem. An ANN is configured for a specific application, such as pattern recognition or data classification, through a learning process.

An ANN consists of three parts or layers: The input layer, a hidden layer and the output layer.  

\begin{figure}[htp]
\centering
\includegraphics[width=4.5cm]{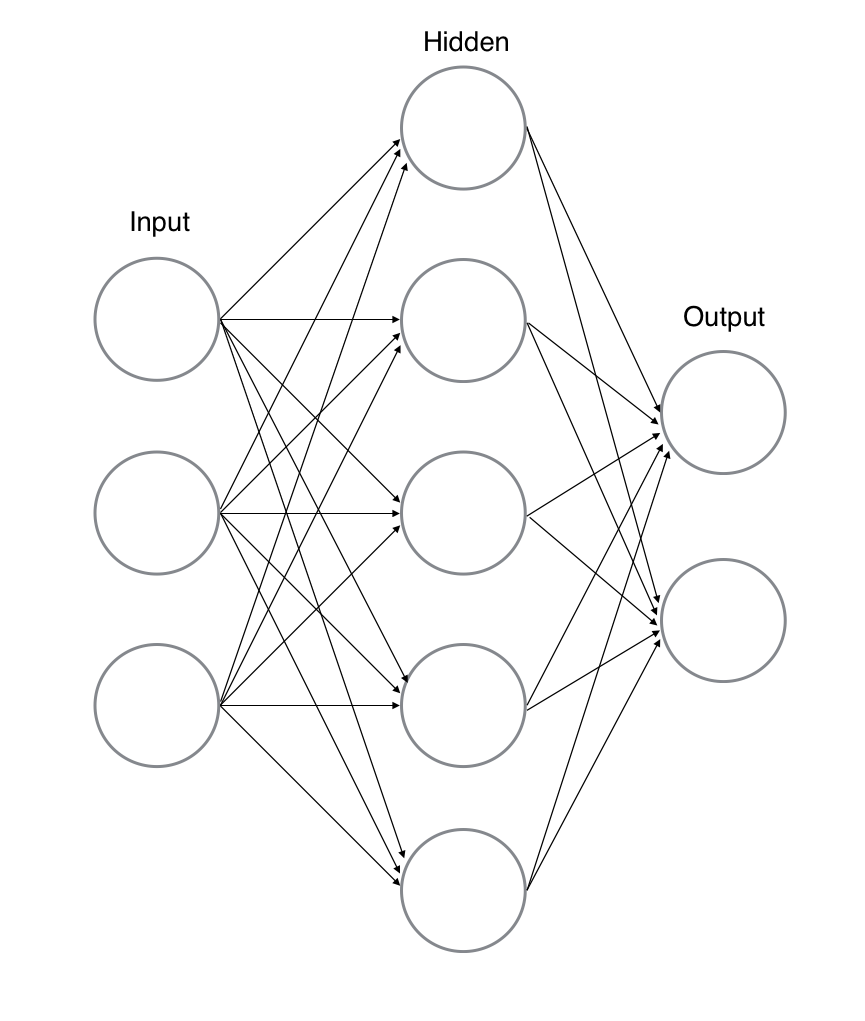}
\caption{Basic structure of an ANN}
\label{fig:Figure1}
\end{figure}

Convolutional Neural Networks are very similar to ordinary Neural Networks. They are also made up of neurons that have learnable weights and biases. The main difference is the number of layers. CNN are just several layers of convolutions with nonlinear activation functions applied to the results. In a traditional NN each input neuron is connected to each output neuron in the next layer. That is called a fully connected layer. In CNNs, instead, convolutions are used over the input layer to compute the output. This results in local connections, where each region of the input is connected to a neuron in the output. Each layer applies different filters, typically hundreds or thousands and combines their results. 

\begin{figure*}[htp]
\centering
\includegraphics[width=14cm]{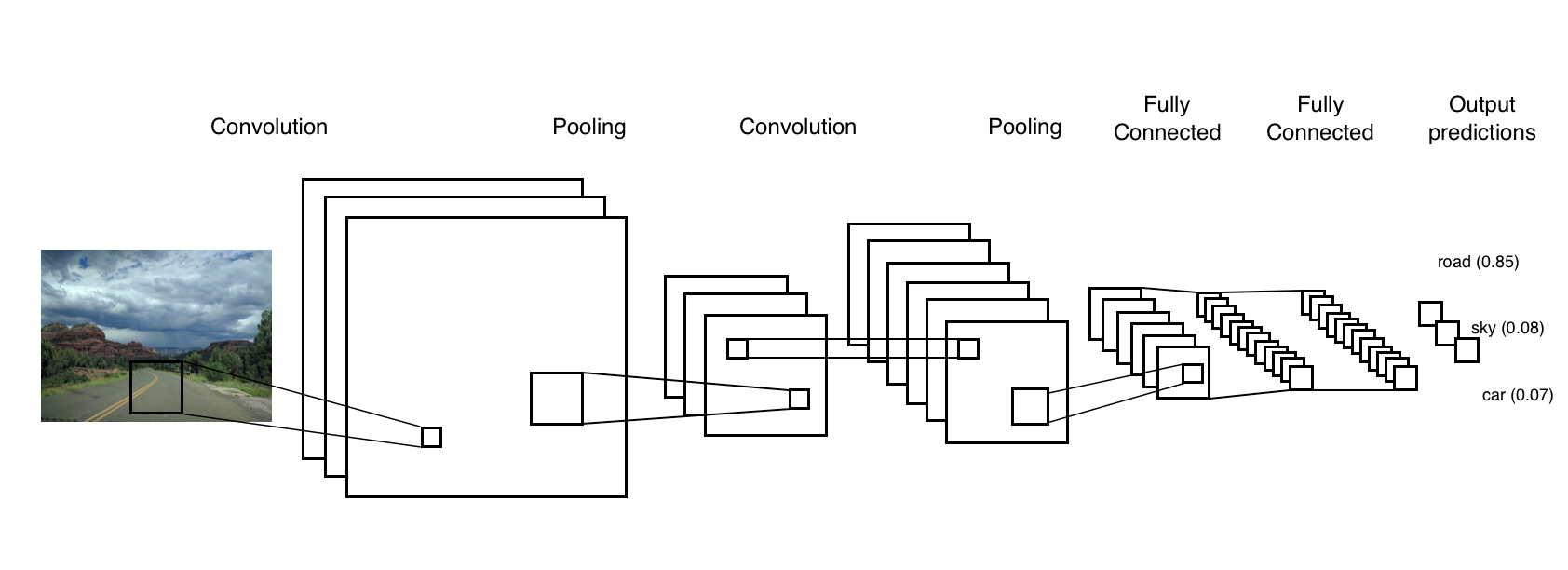}
\caption{Basic structure of a CNN}
\label{fig:Figure2}
\end{figure*}

A key aspect of Convolutional Neural Networks is the use of pooling layers, typically applied after the convolutional layers. Pooling layers subsample their input. The most common way to perform pooling it to apply a max operation to the result of each filter. The pooling process can also be applied over a window. There are two main reasons to perform pooling.

One property of pooling is that it provides a fixed size output matrix, which typically is required for classification. This allows the use of variable size sentences, and variable size filters, but always obtaining the same output dimensions to feed into a classifier.

Pooling also reduces the output dimensionality while keeping the most salient information. You can think of each filter as detecting a specific feature. If this feature occurs somewhere in the sentence, the result of applying the filter to that region will yield a large value, but a small value in other regions. By performing the max operation information is kept about whether or not the feature appeared in the sentence, but information is lost about where exactly it appeared. Resuming, global information about locality is lost (where in a sentence something happens), but local information is kept since it is captured by the filters.

During the training phase, a CNN automatically learns the values of its filters based on the task that to be performed. For example, in Image Classification a CNN may learn to detect edges from raw pixels in the first layer, then use the edges to detect simple shapes in the second layer, and then use these shapes to deter higher-level features, such as facial shapes in higher layers. The last layer is then a classifier that uses these high-level features.

Instead of image pixels, the input to most NLP tasks are sentences or documents represented as a matrix. Each row of the matrix corresponds to one token, typically a word, but it could be a character. That is, each row is vector that represents a word. Typically, these vectors are word embeddings (low-dimensional representations), but they could also be one-hot vectors that index the word into a vocabulary. For a 10 word sentence using a 100-dimensional embedding we would have a 10x100 matrix as our input. 
\begin{figure*}[htp]
\centering
\includegraphics[width=12cm]{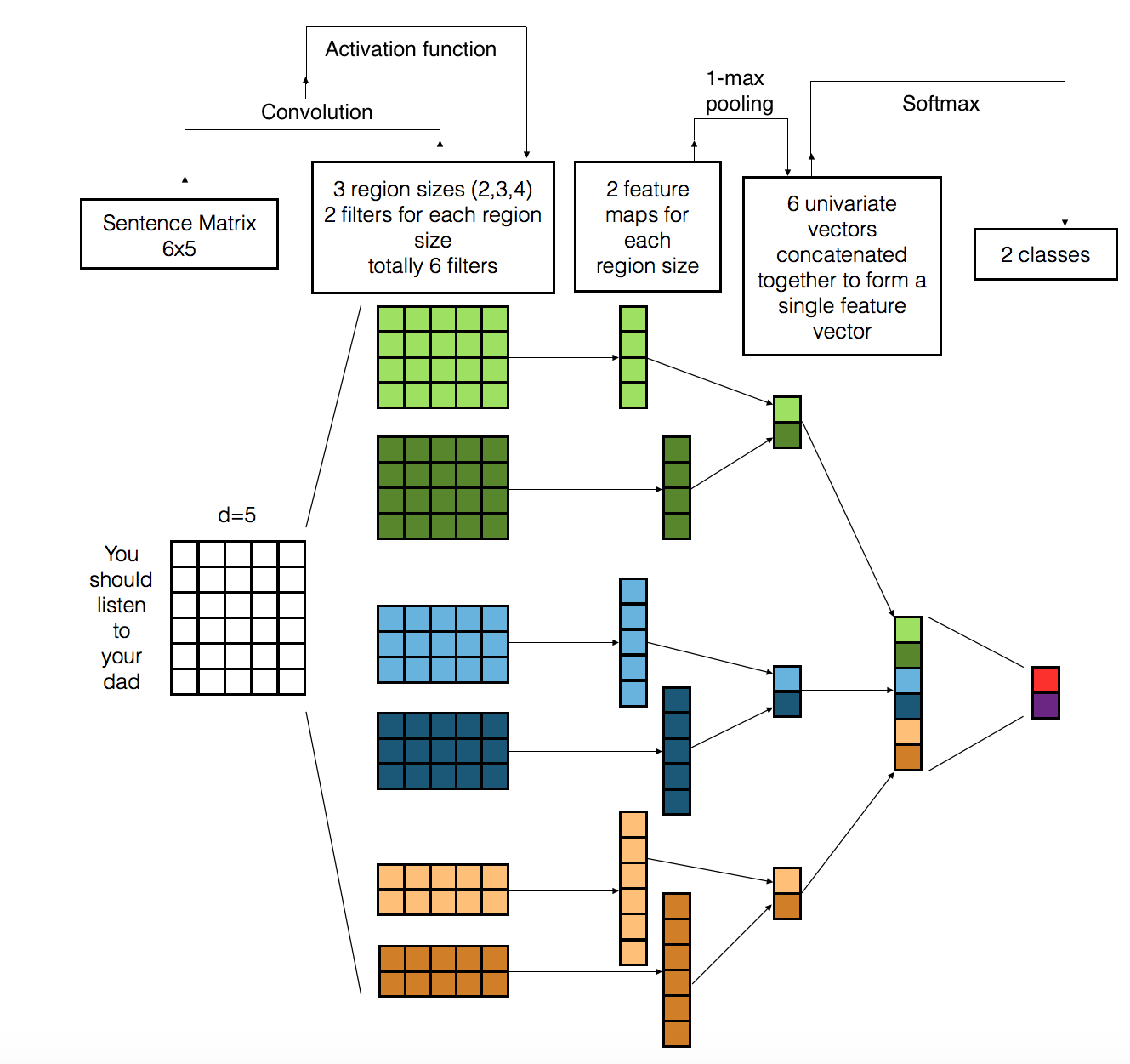}
\caption{How the CNN works}
\label{fig:Figure3}
\end{figure*}

In computer vision, the filters slide over local patches of an image, but in NLP filters slide over full rows of the matrix (words). Thus, the width of the filters is usually the same as the width of the input matrix. The height, or region size, may vary, but sliding windows over 2-5 words at a time is the typical size.

\section{Motivation}
In this paper, Bitvai et al. compare the efficiency of an CNN over an ANN. They consider problem of predicting the future box-office takings of movies based on reviews by movie critics and movie attributes. 
An artificial neural network (ANN) is proposed for modelling text regression. In language processing, ANNs were first proposed for probabilistic language modelling, followed by models of sentences and parsing inter alia. These approaches have shown strong results through automatic learning dense low-dimensional distributed representations for words and other linguistic units, which have been shown to encode important aspects of language syntax and semantics. They also develop a convolutional neural network, inspired by their breakthrough results in image processing and recent applications to language processing. Past works have mainly focused on ?big data? problems with plentiful training examples. Given the large numbers of parameters, often in the millions, one would expect that such models can only be effectively learned on very large datasets. However in this paper they show that a complex deep convolution network can be trained on about a thousand training examples, although careful model design and regularisation is paramount.
They consider the problem of predicting the future box-office takings of movies based on reviews by movie critics and movie attributes. Their approach is based on the method and dataset of Joshi et al. (2010), who presented a linear regression model over uni-, bi-, and tri-gram term frequency counts extracted from reviews, as well as movie and reviewer metadata. This problem is especially interesting, as comparatively few instances are available for training while each instance (movie) includes a rich array of data including the text of several critic reviews from various review sites, as well as structured data (genre, rating, actors, etc.) Inspired by Joshi et al. (2010) their model also operates over n-grams, 1 ? n ? 3, and movie metadata, using an ANN instead of a linear model. They use word embeddings to represent words in a low dimensional space, a convolutional network with max-pooling to represent documents in terms of n-grams, and several fully connected hidden layers to allow for learning of complex non-linear interactions. They show that including non-linearities in the model is crucial for accurate modelling, providing a relative error reduction of 40 per cent (MAE) over the best linear model. Their final contribution is a novel means of model interpretation. 

Although it is notoriously difficult to interpret the parameters of an ANN, they show a simple method of quantifying the effect of text n-grams on the prediction output. This allows for identification of the most important textual inputs, and investigation of non-linear interactions between these words and phrases in different data instances.
\section{Types of Deep Neural Networks}
\subsection{Recurrent neural network}
The idea behind RNNs is to make use of sequential information. In a traditional neural network all inputs (and outputs) are independent of each other. But for many tasks that results in a bad performance. If the next word in a sentence is going to be predicted, there is the need know which words came before it. RNNs are called recurrent because they perform the same task for every element of a sequence, with the output being depended on the previous computations. Another way to think about RNNs is that they have a memory which captures information about what has been calculated so far. Theoretically RNNs can make use of information in arbitrarily long sequences, but in practice they are limited to looking back only a few steps. In Figure 4 we can see what a typical RNN looks like.

\begin{figure}[htp]
\centering
\includegraphics[width=8cm]{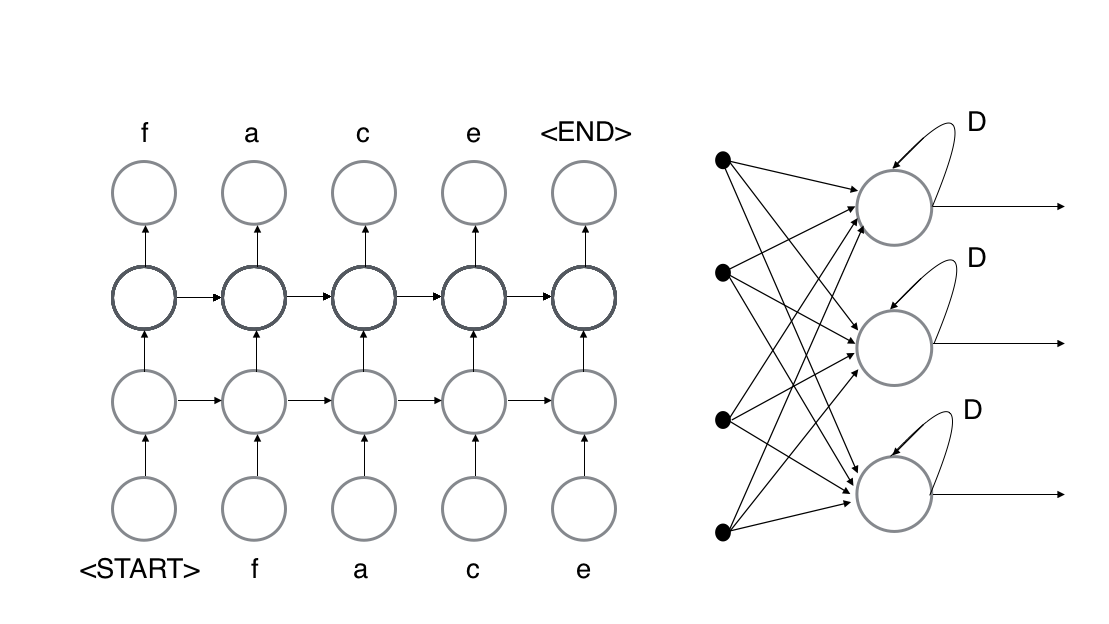}
\caption{Recurrent Network}
\label{fig:Figure4}
\end{figure}

Over the years researchers have developed more sophisticated types of RNNs to deal with some of the shortcomings of the original RNN model. 

\subsubsection{Bidirectional RNN}
Bidirectional RNNs are based on the idea that the output at time t may not only depend on the previous elements in the sequence, but also future elements. For example, to predict a missing word in a sequence you want to look at both the left and the right context. Bidirectional RNNs are quite simple. They are just two RNNs stacked on top of each other. The output is then computed based on the hidden state of both RNNs.

\begin{figure}[htp]
\centering
\includegraphics[width=6cm]{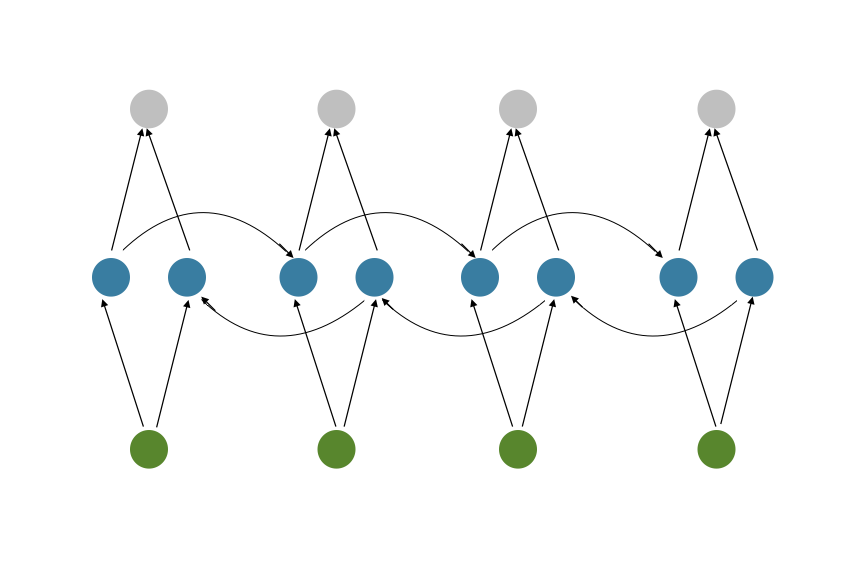}
\caption{Structure of a bidirectional RNN}
\label{fig:Figure5}
\end{figure}

\subsubsection{Deep RNN}
Deep (Bidirectional) RNNs are similar to Bidirectional RNNs, only that we now have multiple layers per time step. In practice this gives us a higher learning capacity (but we also need a lot of training data).

\begin{figure}[htp]
\centering
\includegraphics[width=6cm]{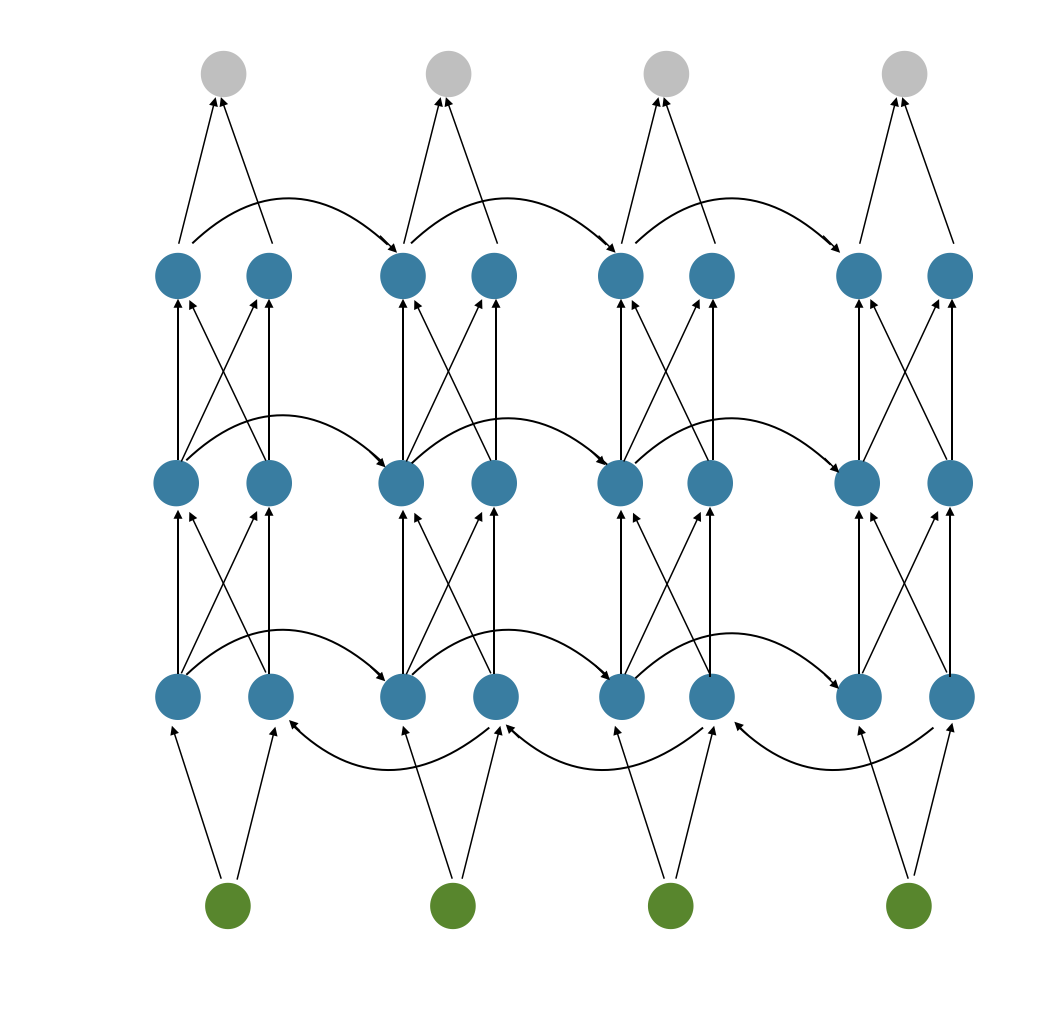}
\caption{Structure of a deep RNN}
\label{fig:Figure6}
\end{figure}

\subsubsection{LSTM networks}
LSTMs don?t have a fundamentally different architecture from RNNs, but they use a different function to compute the hidden state. The memory in LSTMs are called cells and you can think of them as black boxes that take as input the previous state and the current input. Internally these cells  decide what to keep in (and what to erase from) memory. They then combine the previous state, the current memory, and the input. It turns out that these types of units are very efficient at capturing long-term dependencies.

\begin{figure}[htp]
\centering
\includegraphics[width=9cm]{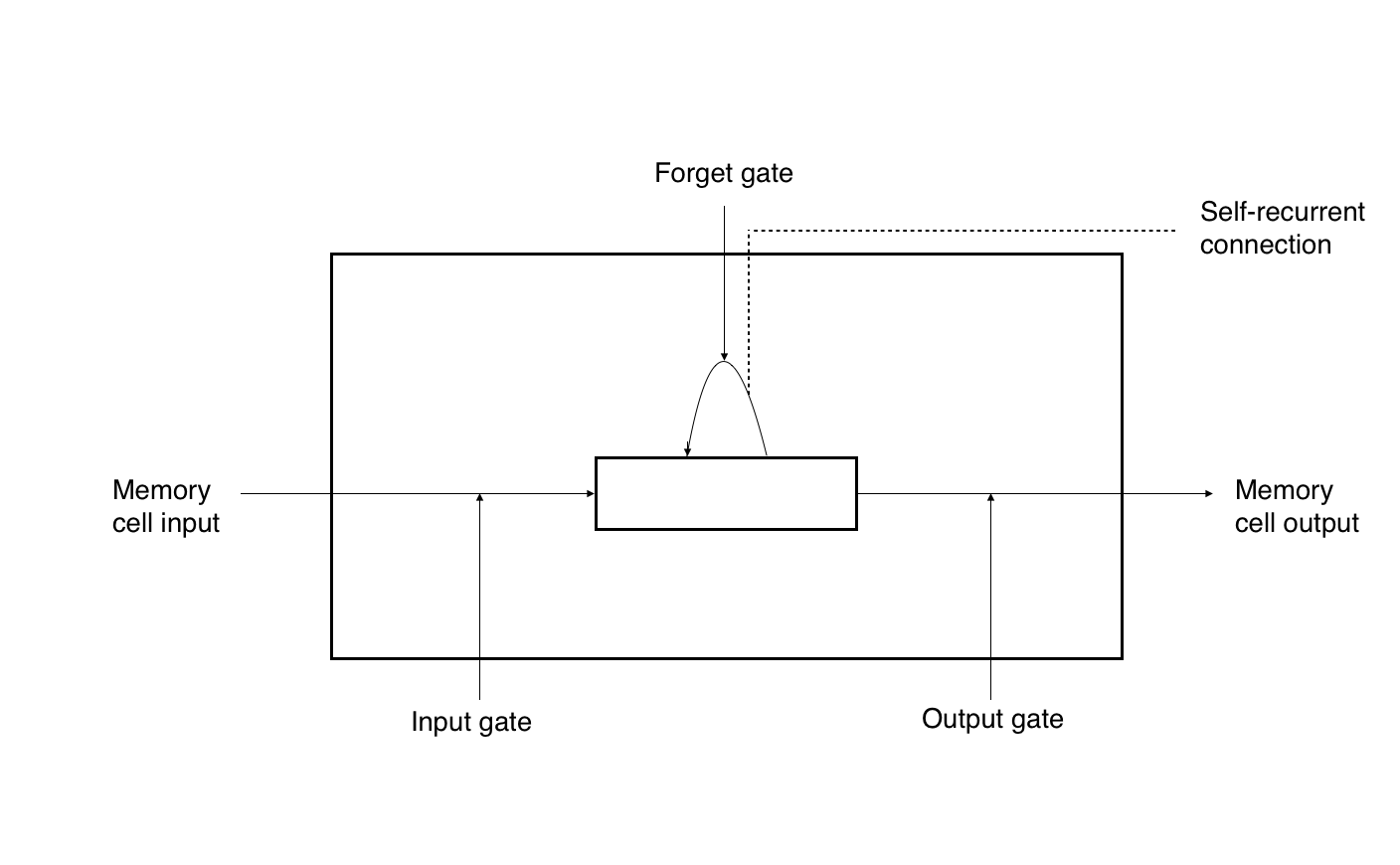}
\caption{Structure of a LSTM}
\label{fig:Figure7}
\end{figure}

\subsection{Recursive neural network}
A recursive neural network (RNN or RCNN) is a deep neural network created by applying the same set of weights recursively over a structure, to produce a structured prediction over the input, or a scalar prediction on it, by traversing a given structure in topological order. RNNs have been successful in learning sequence and tree structures in natural language processing, mainly phrase and sentence continuous representations based on word embedding.

RNN is a general architecture to model the distributed representations of a phrase or sentence with its dependency tree. It can be regarded as semantic modelling of text sequences and handle the input sequences of varying length into a fixed-length vector. The parameters in RCNN can be learned jointly with some other NLP tasks, such as text classification. 

Each RNN unit can model the complicated interactions of the head word and its children. Combined with a specific task, RNN can capture the most useful semantic and structure information by the convolution and pooling layers. 

Recursive neural networks, comprise a class of architecture that operates on structured inputs, and in particular, on directed acyclic graphs. A recursive neural network can be seen as a generalization of the recurrent neural network, which has a specific type of skewed tree structure. They have been applied to parsing, sentence-level sentiment analysis, and paraphrase detection. Given the structural representation of a sentence, e.g. a parse tree, they recursively generate parent representations in a bottom-up fashion, by combining tokens to produce representations for phrases, eventually producing the whole sentence. The sentence-level representation (or, alternatively, its phrases) can then be used to make a final classification for a given input sentence.

Similar to how recurrent neural networks are deep in time, recursive neural networks are deep in structure, because of the repeated application of recursive connections. Recently, the notions of depth in time the result of recurrent connections, and depth in space the result of stacking multiple layers on top of one another, are distinguished for recurrent neural networks. In order to combine these concepts, deep recurrent networks were proposed. They are constructed by stacking multiple recurrent layers on top of each other, which allows this extra notion of depth to be incorporated into temporal processing. Empirical investigations showed that this results in a natural hierarchy for how the information is processed.
\begin{figure}[htp]
\centering
\includegraphics[width=6cm]{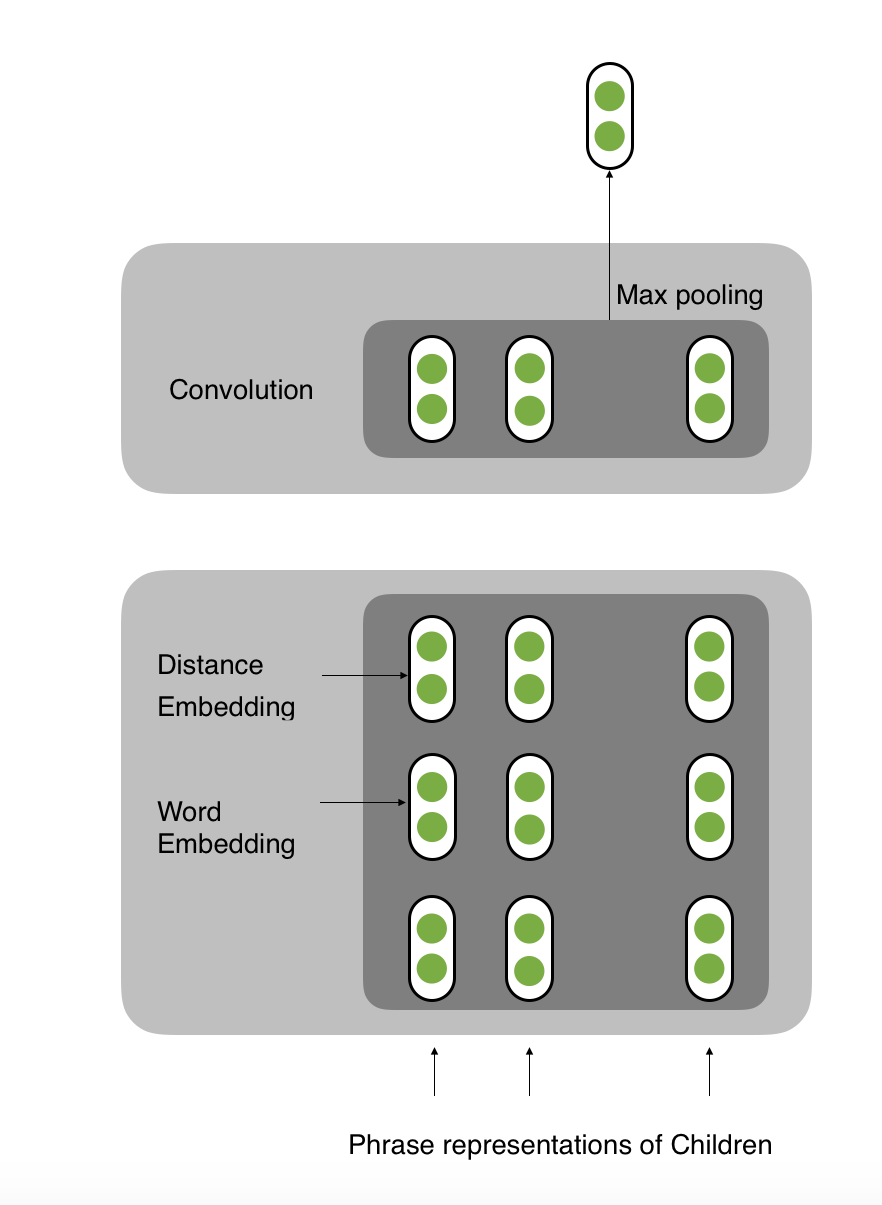}
\caption{Recursive Network}
\label{fig:Figure8}
\end{figure}

\subsection{Dependency based neural network}
In order to capture long-distance dependencies a dependency-based convolution model (DCNN) is proposed.
DCNN consists of a convolutional layer built on top of Long Short-Term Memory (LSTM) networks. DCNN takes slightly different forms depending on its input. For a single sentence, the LSTM network processes the sequence of word embeddings to capture long-distance dependencies within the sentence. The hidden states of the LSTM are extracted to form the low-level representation, and a convolutional layer with variable-size filters and max-pooling operators follows to extract task-specific features for classification purposes. As for document modeling, DCNN first applies independent LSTM networks to each subsentence. Then a second LSTM layer is added between the first LSTM layer and the convolutional layer to encode the dependency across different sentences. 
 
\begin{figure*}[htp]
\centering
\includegraphics[width=14cm]{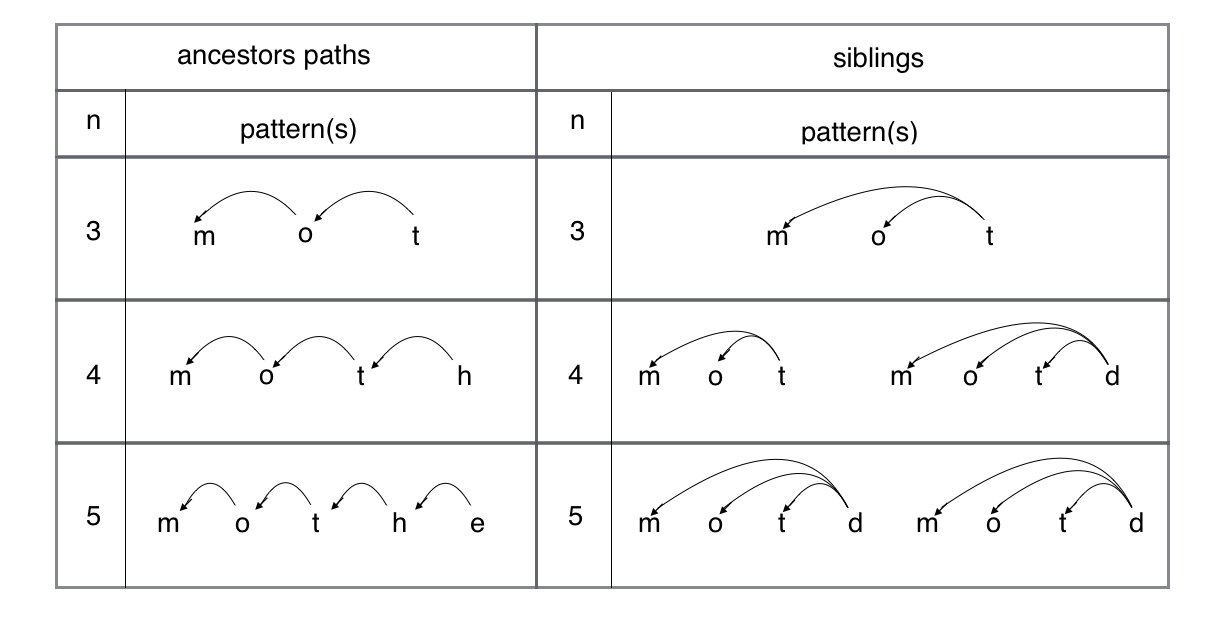}
\caption{Dependency Network}
\label{fig:Figure9}
\end{figure*}

\subsection{Dynamic k-max pooling neural network}
Dynamic k-max pooling is a generalization of the max pooling operator. The max pooling operator is a non-linear subsampling function that returns the maximum of a set of values. The operator is generalized in two respects. First, k-max pooling over a linear sequence of values returns the subsequence of k maximum values in the sequence, instead of the single maximum value. Secondly, the pooling parameter k can be dynamically chosen by making k a function of other aspects of the network or the input.

The convolutional layers apply one-dimensional filters across each row of features in the sentence matrix. Convolving the same filter with the n-gram at every position in the sentence allows the features to be extracted independently of their position in the sentence. A convolutional layer followed by a dynamic pooling layer and a non-linearity form a feature map. Like in the convolutional networks for object recognition (LeCun et al., 1998), the representation is enriched in the first layer by computing multiple feature maps with different filters applied to the input sentence. Subsequent layers also have multiple feature maps computed by convolving filters with all the maps from the layer below. The weights at these layers form an order-4 tensor. The resulting architecture is dubbed a Dynamic Convolutional Neural Network.
Multiple layers of convolutional and dynamic pooling operations induce a structured feature graph over the input sentence. Insert figure. Figure 10 illustrates such a graph. Small filters at higher layers can capture syntactic or semantic relations between noncontinuous phrases that are far apart in the input sentence. The feature graph induces a hierarchical structure somewhat akin to that in a syntactic parse tree. The structure is not tied to purely syntactic relations and is internal to the neural network.
 
\begin{figure*}[htp]
\centering
\includegraphics[width=16cm]{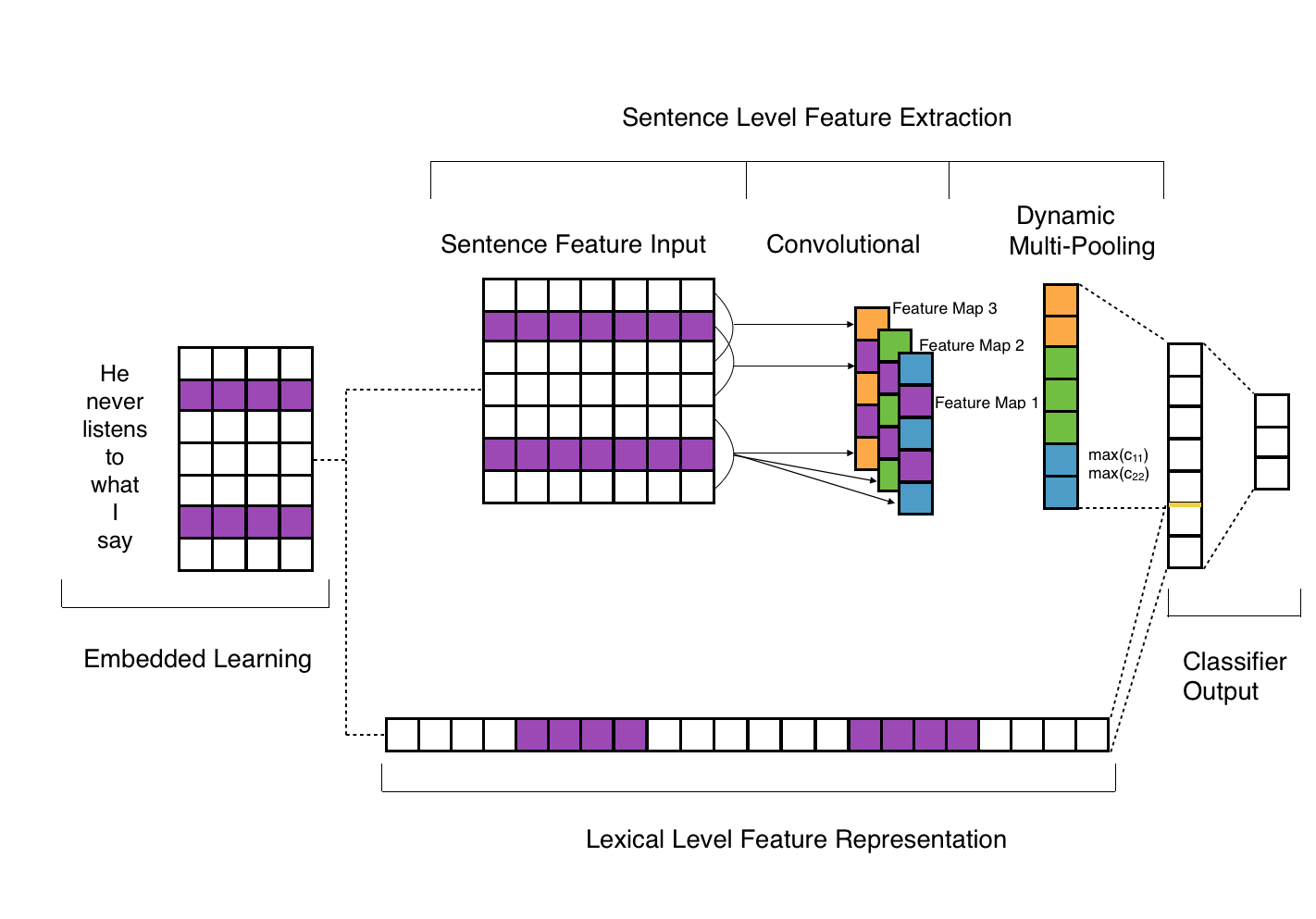}
\caption{Dynamic Multi-pooling Network}
\label{fig:Figure10}
\end{figure*}

\subsection{Other neural networks}
\subsubsection{Multi-column CNN}
This model shares the same word embeddings, and s multiple columns of convolutional neural networks. The number of columns usually used is three, but it can have more or less depending on the context in which it has to be used. These columns are used to analyze different aspects of a question, i.e., answer path, answer context, and answer type. Typically this framework is combined with the learning of embeddings. The overview of this framework is shown in Figure 11. For instance, for the question when did Avatar release in UK, the related nodes of the entity Avatar are queried from FREEBASE. These related nodes are regarded as candidate answers (Cq). Then, for every candidate answer a, the model predicts a score S (q, a) to determine whether it is a correct answer or not.

\begin{figure*}[htp]
\centering
\includegraphics[width=16cm]{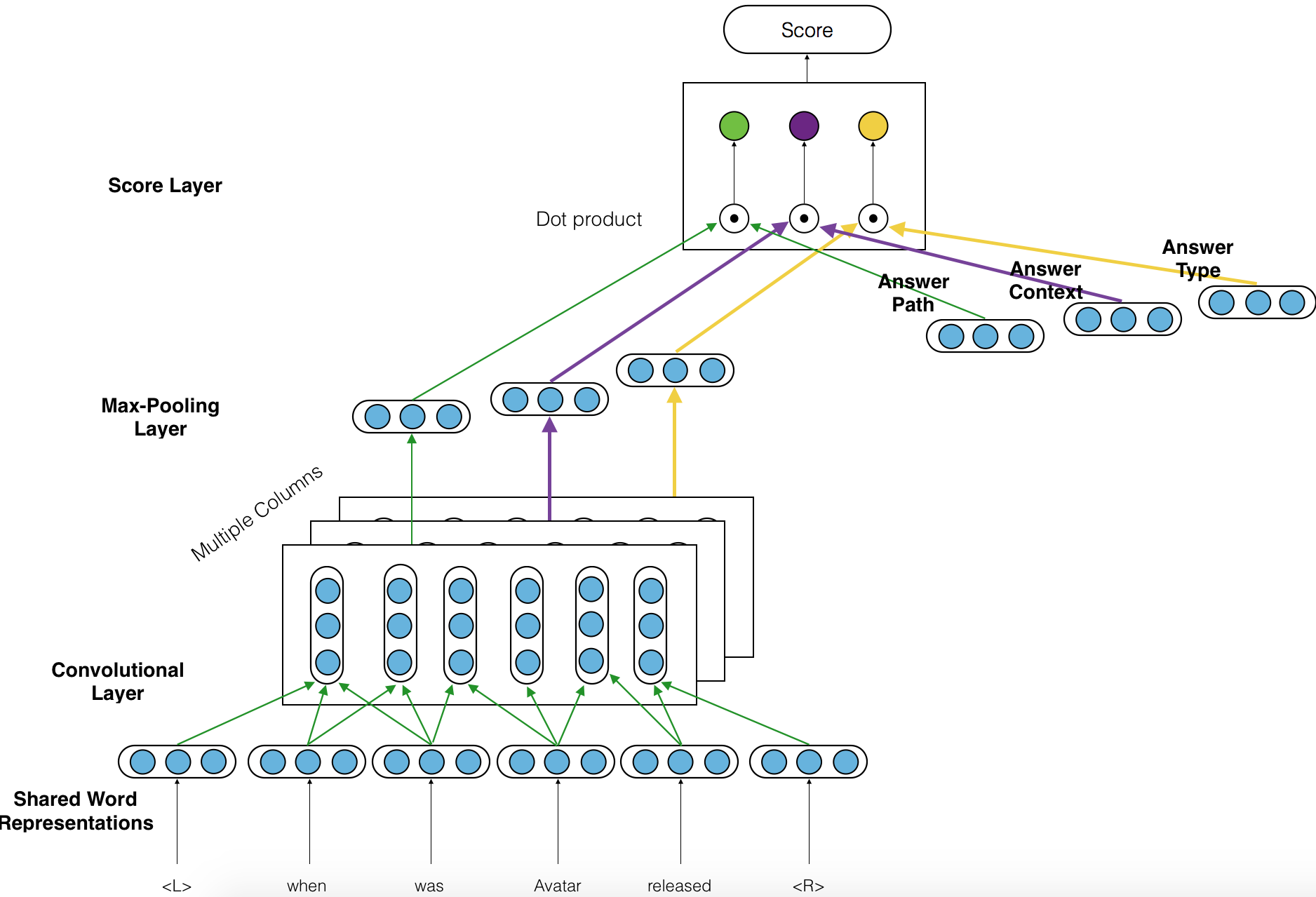}
\caption{Multi-column Network}
\label{fig:Figure11}
\end{figure*}

\subsubsection{Ranking CNN}

\begin{figure}[htp]
\centering
\includegraphics[width=6cm]{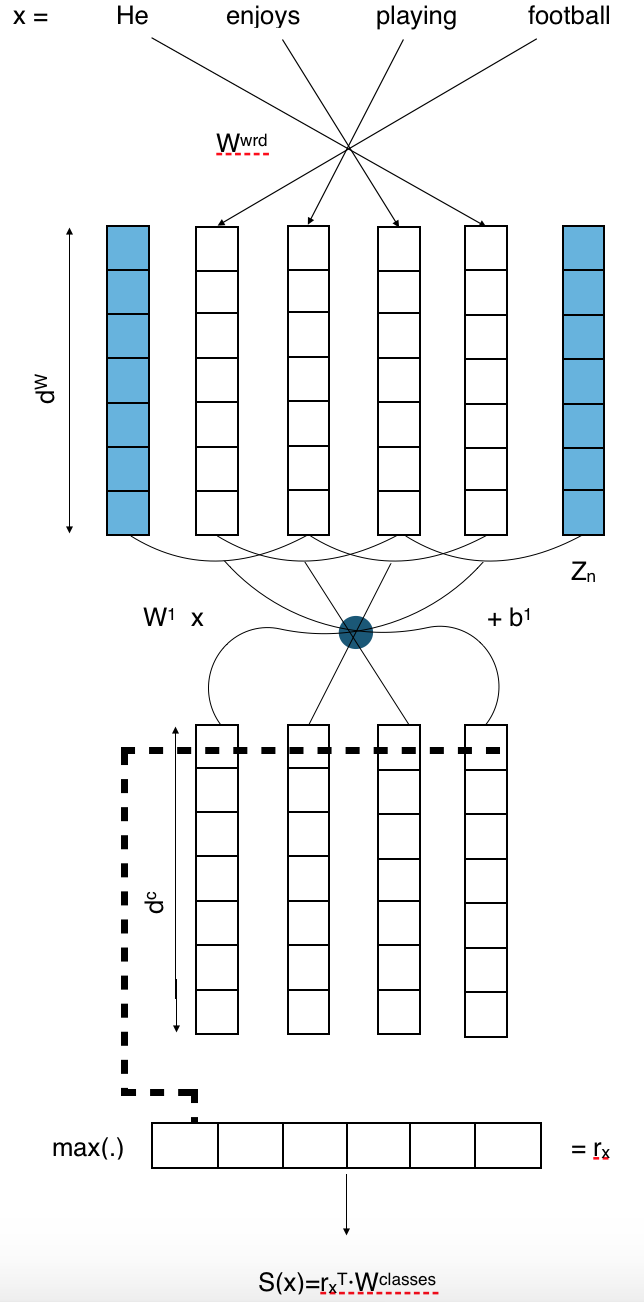}
\caption{Ranking Network}
\label{fig:Figure12}
\end{figure}

\subsubsection{Context dependent CNN}
The model architecture, shown in Figure x, is a variant of the convolutional architecture of Hu et al. (2014). It consists of two components:
? convolutional sentence model that summarizes the meaning of the source sentence and the target phrase;
? matching model that compares the two representations with a multi-layer perceptron (Bengio, 2009).
Let E be a target phrase and F be the source sentence that contains the source phrase aligning to E. First of all F and E are projected into feature vectors x and y via the convolutional sentence model, and then the matching score s(x, y) is computed by the matching model. Finally, the score is introduced into a conventional SMT system as an additional feature. Convolutional sentence model. As shown in Figure 13, the model takes as input the embeddings of words (trained beforehand elsewhere) in F and E. It then iteratively summarizes the meaning of the input through layers of convolution and pooling, until reaching a fixed length vectorial representation in the final layer.
In Layer-1, the convolution layer takes sliding windows on F and E respectively, and models all the possible compositions of neighbouring words. The convolution involves a filter to produce a new feature for each possible composition.

\begin{figure*}[htp]
\centering
\includegraphics[width=14cm]{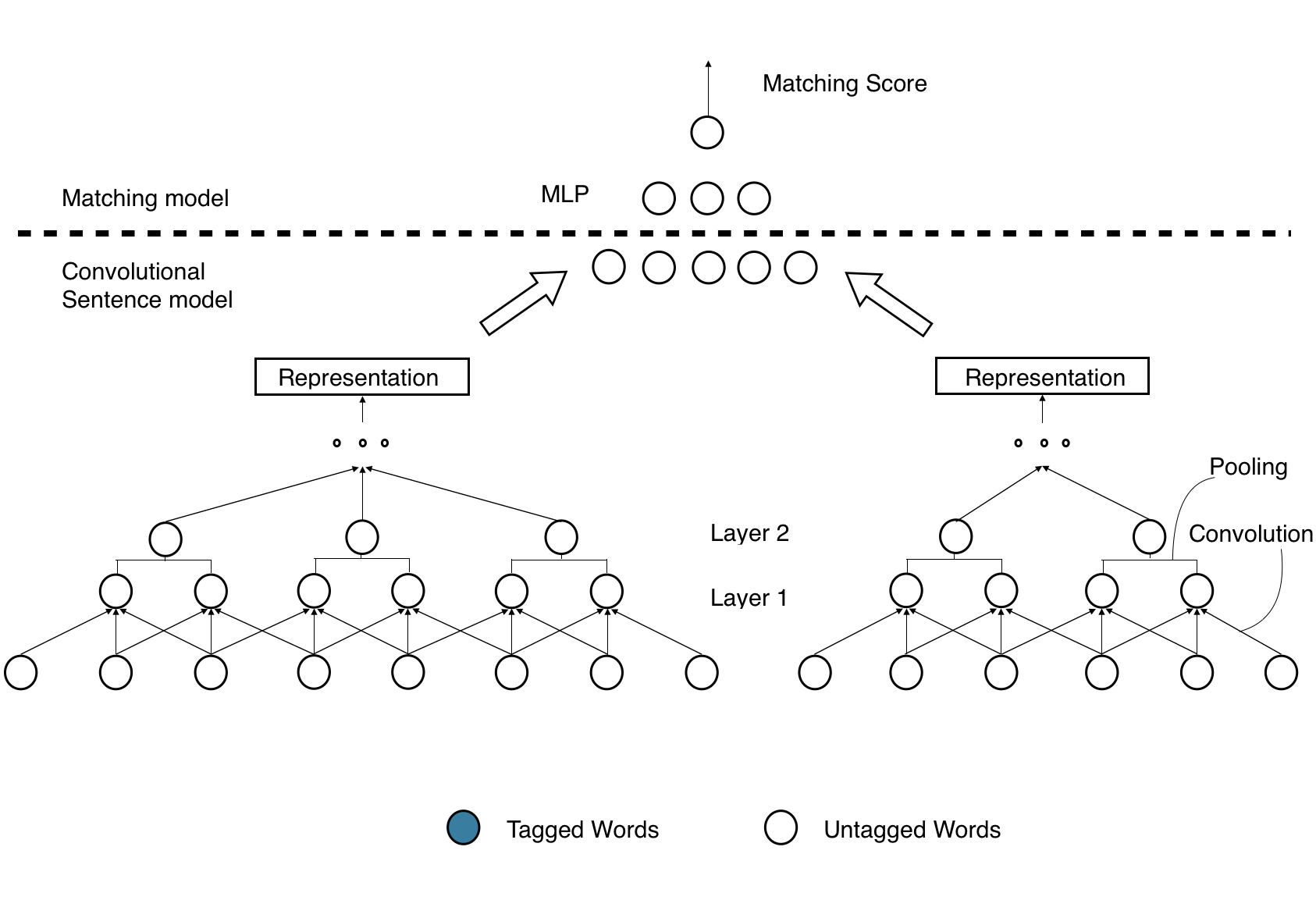}
\caption{Context dependent Network}
\label{fig:Figure13}
\end{figure*}

\section{Natural Language Processing}

\subsection{Basic NLP}

\subsubsection{A Re-ranking Model for Dependency Parser with Recursive Convolutional Neural Network}
In this paper, Zhu et al. propose a recursive convolutional neural network (RCNN) architecture to capture syntactic and compositional-semantic representations of phrases and words. RCNN is a general architecture and can deal with k-ary parsing tree, therefore it is very suitable for dependency parsing. For each node in a given dependency tree, they first use a RCNN unit to model the interactions between it and each of its children and choose the most informative features by a pooling layer. Thus, the RCNN unit can be applied recursively to get the vector representation of the whole dependency tree. The output of each RCNN unit is used as the input of the RCNN unit of its parent node, until it outputs a single fixed-length vector at root node. When applied to the re-ranking model for parsing, RCNN improve the accuracy of base parser to make accurate parsing decisions. The experiments on two benchmark datasets show that RCNN outperforms the state-of-the-art models. The results obtained for this paper can be seen in Table 1.

\begin{table}[ht]
\begin{center}
 \begin{tabular}{||l c||} 
 \hline
   & UAS \\ [0.5ex] 
 \hline\hline
 Traditional Methods &   \\ 
 \hline
 Zhang and Clark (2008) & 91.4 \\
 \hline
 Huang and Sagae (2010) & 92.1 \\
 \hline
 Distributed Representations &   \\
 \hline
 Stenetorp (2013) & 86.25 \\
 \hline
 Chen et al. (2014) & 93.74 \\
 \hline
 Chen and Manning (2014) & 92.0 \\
 \hline
 Re-rankers &   \\
 \hline
 Hayashi et al. (2013) & 93.12 \\
 \hline
 Le and Zuidema (2014) & 93.12 \\
 \hline
 Our baseline & 92.35 \\
 \hline
 Our re-ranker & 93.83(+1.48) \\
 \hline
 Our re-ranker (with oracle) & 94.16 \\ [1ex] 
 \hline
\end{tabular}
\end{center}
\caption{results}
\end{table}

\subsubsection{Semantic Clustering and Convolutional Neural Network for Short Text Categorization}
In this paper, Wang et al. propose a novel method to model short texts based on semantic clustering and convolutional neural network. Particularly, they first discover semantic cliques in embedding spaces by a fast clustering algorithm: (1) semantic cliques are discovered using fast clustering method based on searching density peaks; (2) for fine-tuning multi- scale SUs, the semantic cliques are used to super- vise the selection stage. Since the neighbors of each word are semantically related in embedding space, clustering methods can be used to discover semantic cliques. Then, multi-scale semantic units are detected under the supervision of semantic cliques, which introduce useful external knowledge for short texts. These meaningful semantic units are combined and fed into convolutional layer, followed by max-pooling operation.

\subsubsection{Capturing Semantic Similarity for Entity Linking with Convolutional Neural Networks}
In this work, Francis-landau et al. present a model that uses convolutional neural networks to capture semantic correspondence between a mention?s context and a proposed target entity. These convolutional networks operate at multiple granularities to exploit various kinds of topic information, and their rich parameterization gives them the capacity to learn which n-grams characterize different topics. They model semantic similarity between a mention's source document context and its potential entity targets using CNNs. CNNs have been shown to be effective for sentence classification tasks and for capturing similarity in models for entity linking so they are expected to be effective at isolating the relevant topic semantics for entity linking. They show that convolutions over multiple granularities of the input document are useful for providing different notions of semantic context. Finally, they show how to integrate these networks with a preexisting entity linking system. Through a combination of these two distinct methods into a single system that leverages their complementary strengths, they achieve state-of-the-art performance across several datasets. The results obtained for this paper can be seen in Table 2. 

\begin{table}[ht]
\begin{center}
 \begin{tabular}{||l c c c||} 
 \hline
   & ACE & CoNLL & WP \\ [0.5ex] 
 \hline\hline
 Google News & 87.5 & 89.6 & 83.8 \\
 \hline
 Wikipedia & 89.5 & 90.6 & 85.5 \\ [1ex] 
 \hline
\end{tabular}
\end{center}
\caption{results}
\end{table}

\subsubsection{Dependency Sensitive Convolutional Neural Networks for Modeling Sentences and Documents}
In this work, Zhang et al. present Dependency Sensitive Convolutional Neural Networks (DSCNN) as a general-purpose classification system for both sentences and documents. DSCNN hierarchically builds textual representations by processing pretrained word embeddings via Long Short-Term Memory networks and subsequently extracting features with convolution operators. Compared with existing recursive neural models with tree structures,DSCNN does not rely on parsers and expensive phrase labeling, and thus is not restricted to sentence-level tasks. Moreover, unlike other CNN-based models that analyze sentences locally by sliding windows, their system captures both the dependency information within each sentence and relationships across sentences in the same document. 
They propose Dependency Sensitive Convolutional Neural Networks (DSCNN), an end-to-end classification system that hierarchically builds textual representations with only root-level labels. 
They evaluate DSCNN on several sentence-level and document-level tasks including sentiment analysis, question type classification, and subjectivity classification. 

\subsection{Information Extraction}

\subsubsection{Event Extraction via Dynamic Multi-Pooling Convolutional Neural Networks}
In this paper, Chen et al. introduce a word-representation model to capture meaningful semantic regularities for words and adopt a framework based on a multi-pooling CNN to capture sentence-level clues. Since CNN can only capture the most important information in a sentence and may miss valuable facts when considering multiple-event sentences, they propose a dynamic multi-pooling convolutional neural network (DMCNN), as seen in CNN type 3. DMNCC uses a dynamic multi-pooling layer according to event triggers and arguments, to reserve more crucial information. Explain a bit more. 

\subsubsection{Event Detection and Domain Adaptation with Convolutional Neural Networks}
In this paper, Nguyen et al. present a convolutional neural network for event detection that automatically learns features from sentences, and minimizes the dependence on supervised toolkits and resources for features, thus alleviating the error propagation and improving the performance for this task. First, they evaluate CNNs for event detection in the general setting and show that CNNs, though not requiring complicated feature engineering, can still outperform the state-of-the-art feature-based methods extensively relying on the other supervised modules and manual resources for features. Second, they investigate CNNs in a domain adaptation (DA) setting for event detection. They demonstrate that CNNs significantly outperform the traditional feature-based methods with respect to generalization performance across domains due to: (i) their capacity to mitigate the error propagation from the preprocessing modules for features, and (ii) the use of word embeddings to induce a more general representation for trigger candidates. 

\subsubsection{Combining Recurrent and Convolutional Neural Networks for Relation Classification}
In this paper Vu et al. present three different approaches. First of all, a new context representation for convolutional neural networks for relation classification (extended middle context). Secondly, they propose connectionist bi-directional recurrent neural networks and introduce ranking loss for their optimization. Finally, they show that combining convolutional and recurrent neural net- works using a simple voting scheme is accurate enough to improve results. 
1) The presented extended middle context, a new context representation for CNNs for relation classification. The extended middle context uses all parts of the sentence (the relation arguments, left of the relation arguments, between the arguments, right of the arguments) and pays special attention to the middle part. 
2) They present connectionist bi-directional RNN models which are especially suited for sentence classification tasks since they combine all intermediate hidden layers for their final decision. Furthermore, the ranking loss function is introduced for the RNN model optimization which has not been investigated in the literature for relation classification before. 
3) Finally, they combine CNNs and RNNs using a simple voting scheme and achieve new state-of-the-art results on the SemEval 2010 benchmark dataset. 
The results obtained for this paper can be seen in Table 3. 

\begin{table}[ht]
\begin{center}
 \begin{tabular}{||l l||} 
 \hline
 Classifier & F1 \\ [0.5ex] 
 \hline\hline
 SVM (Rink and Harabagiu, 2010b) & 82.2 \\ 
 \hline
 RNN (Socher et al., 2012) & 77.6 \\
 \hline
 MVRNN (Socher et al., 2012) & 82.4 \\
 \hline
 CNN (Zeng et al., 2014) & 82.7 \\
 \hline
 FCM (Yu et al., 2014) & 83.0 \\
 \hline
 bi-RNN (Zhang and Wang, 2015) & 82.5 \\
 \hline
 CR-CNN (Dos Santos et al., 2015) & 84.1 \\
 \hline
 R-CNN & 83.4 \\
 \hline
 ER-CNN & 84.2 \\
 \hline
 ER-CNN+R-RNN & 84.9 \\ [1ex] 
 \hline
\end{tabular}
\end{center}
\caption{results}
\end{table}

\subsubsection{Comparing Convolutional Neural Networks to Traditional Models for Slot Filling}
In this paper Adel et al. address relation classification in the context of slot filling, the task of finding and evaluating fillers for different slots. They investigate three complementary approaches to relation classification.
The first approach is pattern matching, a leading approach in the TAC evaluations. Fillers are validated based on patterns. In this work, they consider patterns learned with distant supervision. Their second approach is support vector machines. Their third approach is a convolutional neural network (CNN). CNN can recognize phrase patterns independent of their position in the sentence. Furthermore, they make use of word embeddings that directly reflect word similarity.
1) They investigate the complementary strengths and weaknesses of different approaches to relation classification and show that their combination can better deal with a diverse set of problems that slot filling poses than each of the approaches individually. 
2) They propose to split the context at the relation arguments before passing it to the CNN in order to better deal with the special characteristics of a sentence in relation classification. This outperforms the state-of-the-art piecewise CNN. 
3) They analyze the effect of genre on slot filling and show that it is an important conflating variable that needs to be carefully examined in research on slot filling. 
4) They provide a benchmark for slot filling relation classification that will facilitate direct comparisons of models in the future and show that results on this dataset are correlated with end-to-end system results. 

\subsection{Summarization}

\subsubsection{Modelling, Visualising and Summarising Documents with a Single Convolutional Neural Network}
In this paper Denil et al. introduce a model that is able to represent the meaning of documents by embedding them in a low dimensional vector space, while preserving distinctions of word and sentence order crucial for capturing nuanced semantics. Their model is based on an extended Dynamic Convolution Neural Network, which learns convolution filters at both the sentence and document level, hierarchically learning to capture and compose low level lexical features into high level semantic concepts. Their model is compositional; it combines word embeddings into sentence embeddings and then further combines the sentence embeddings into document embeddings. This means that their model is divided into two levels, a sentence level and a document level, both of which are implemented using CNN. At the sentence level CNN are used to transform embeddings for the words in each sentence into an embedding for the entire sentence. At the document level another CNN is used to transform sentence embeddings from the first level into a single embedding vector that represents the entire document. Since their model is based on convolutions, it is able to preserve ordering information between words in a sentence and between sentences in a document. The results obtained for this paper can be seen in Table 4.

\begin{table}[ht]
\begin{center}
 \begin{tabular}{||l l||} 
 \hline
 Model & Accuracy \\ [0.5ex] 
 \hline\hline
 BoW & 88.23 \\ 
 \hline
 Full+BoW & 88.33 \\
 \hline
 Full+Unlabelled+BoW & 88.89 \\
 \hline
 WRRBM & 87.42 \\
 \hline
 WRRBM+BoW (bnc) & 89.23 \\
 \hline
 SVM-bi & 86.95 \\
 \hline
 NBSVM-uni & 88.29 \\
 \hline
 NBSVM-bi & 91.22 \\
 \hline
 Paragraph Vector & 92.58 \\
 \hline
 Their model & 89.38 \\ [1ex] 
 \hline
\end{tabular}
\end{center}
\caption{results}
\end{table}

\subsection{Machine Translation}

\subsubsection{Context-Dependent Translation Selection Using Convolutional Neural Network}
In this paper, Hu et al. propose a novel method for translation selection in statistical machine translation, in which a convolutional neural network is employed to judge the similarity between a phrase pair in two languages. The specifically designed convolutional architecture encodes not only the semantic similarity of the translation pair, but also the context containing the phrase in the source language. Therefore, their approach is able to capture context-dependent semantic similarities of translation pairs. A curriculum learning strategy is adopted to train the model: the training examples are classified into easy, medium, and difficult categories, and gradually build the ability of representing phrases and sentence-level contexts by using training examples from easy to difficult. 

\subsubsection{Encoding Source Language with Convolutional Neural Network for Machine Translation}
In this paper, Meng et al. use a CNN plus gating approach. They give a more systematic treatment by summarizing the relevant source information through a convolutional architecture guided by the target information. With different guiding signals during decoding, their specifically designed convolution+gating architectures can pinpoint the parts of a source sentence that are relevant to predicting a target word, and fuse them with the context of entire source sentence to form a unified representation. This representation, together with target language words, are fed to a deep neural network (DNN) to form a stronger neural network joint model,NNJM. Experiments on two NIST Chinese-English translation tasks show that the proposed model can achieve significant improvements over the previous NNJM. The results obtained for this paper can be seen in Table 5 and Table 6.

\begin{table}[ht]
\begin{center}
 \begin{tabular}{||c c c c||} 
 \hline
 Systems & MT04 & MT05 & Average \\ [0.5ex] 
 \hline\hline
 Deep2str & 34.89 & 32.24 & 33.57 \\ 
 \hline
 tagCNN & 36.33 & 33.37 & 34.85 \\
 \hline
 tagCNN-dep & 36.53 & 33.61 & 35.08 \\ [1ex] 
 \hline
\end{tabular}
\end{center}
\caption{results}
\end{table}

\begin{table}[ht]
\begin{center}
 \begin{tabular}{||c c c c||} 
 \hline
 Systems & MT04 & MT05 & Average \\ [0.5ex] 
 \hline\hline
 Deep2str & 34.89 & 32.24 & 33.57 \\ 
 \hline
 inCNN & 36.92 & 33.72 & 35.32 \\
 \hline
 inCNN-2pooling & 36.33 & 32.88 & 34.61 \\
 \hline
 inCNN-4pooling & 36.46 & 33.01 & 34.74 \\
 \hline
 inCNN-8pooling & 36.57 & 33.39 & 34.98 \\ [1ex] 
 \hline
\end{tabular}
\end{center}
\caption{results}
\end{table}

\subsection{Question Answering}

\subsubsection{Question Answering over Freebase with Multi-Column Convolutional Neural Networks}
In this paper, Dong et al. introduce the multi-column convolutional neural networks (MCCNNs) to automatically analyze questions from multiple aspects. Specifically, the model shares the same word embeddings to represent question words. MCCNNs use different column networks to extract answer types, relations, and context information from the input questions. The entities and relations in the knowledge base are also represented as low-dimensional vectors. Then, a score layer is employed to rank candidate answers according to the representations of questions and candidate answers. Their proposed information extraction based method utilizes question-answer pairs to automatically learn the model without relying on manually annotated logical forms and hand-crafted features. They do not use any pre-defined lexical triggers and rules. In addition, the question paraphrases are also used to train networks and generalize for the unseen words in a multi-task learning manner. The results obtained for this paper can be seen in Table 7.

\begin{table}[ht]
\begin{center}
 \begin{tabular}{||l c c||} 
 \hline
 Method & F1 & P@1 \\ [0.5ex] 
 \hline\hline
 (Berant et al., 2013) & 31.4 & - \\ 
 \hline
 (Berant and Liang, 2014) & 39.9 & - \\
 \hline
 (Bao et al., 2014) & 37.5 & - \\
 \hline
 (Yao and Van Durme, 2014) & 33.0 & - \\
 \hline
 (Bordes et al., 2014a) & 39.2 & 40.4 \\
 \hline
 (Bordes et al., 2014b) & 28.7 & 31.3 \\
 \hline
 MCCNN (theirs) & 40.8 & 45.1 \\ [1ex] 
 \hline
\end{tabular}
\end{center}
\caption{results}
\end{table}

\subsubsection{Modeling Relational Information in Question-Answer Pairs with Convolutional Neural Networks}
In this paper, Severyn et al. propose convolutional neural networks for learning an optimal representation of question and answer sentences. The main aspect of this work is the use of relational information given by the matches between words from the two members of the pair. The matches are encoded as embeddings with additional parameters (dimensions), which are tuned by the network. These allows for better capturing interactions between questions and answers, resulting in a significant boost in accuracy.
The distinctive properties of their model are:
1) State-of-the-art use of distributional sentence model for learning to map input sentences to vectors, which are then used to measure the similarity between them.
2) Their model encodes question-answer pairs in a richer representation using not only their similarity score but also their intermediate representations.
3) They augment the word embeddings with additional dimensions to encode the fact that certain words overlap in a given question-answer pair and let the network tune these parameters. 
4) The architecture of our net- work makes it straightforward to include any additional features encoding question-answer similarities
5) Finally their model is trained end-to-end starting from the input sentences to producing a final score that is used to rerank answers. They only require to initialize word embeddings trained on some large unsupervised corpora. However, given a large training set the network can also optimize the embeddings directly for the task, thus omitting the need for pre-training of the word embeddings. The results obtained for this paper can be seen in Table 8.

\begin{table}[ht]
\begin{center}
 \begin{tabular}{||l c c||} 
 \hline
 Model & MAP & MRR \\ [0.5ex] 
 \hline\hline
 Wang et al. (2007) & .6029 & .6852 \\ 
 \hline
 Heilman and Smith (2010) & .6091 & .6917 \\
 \hline
 Wang and Manning (2010) & .5951 & .6951 \\
 \hline
 Yao et al. (2013) & .6307 & .7477 \\
 \hline
 Severyn and Moschitti (2013) & .6781 & .7358 \\
 \hline
 Yih et al. (2013) & .7092 & .7700 \\
 \hline
 Yu et al. (2014) & .7113 & .7846 \\
 \hline
 Wang and Ittycheriah (2015) & .7063 & .7740 \\
 \hline
 Yin et al. (2015) & .6951 & .7633 \\
 \hline
 Miao et al. (2015) & .7339 & .8117 \\
 \hline
 CNNR on (TRAIN) & .6857 & .7660 \\
 \hline
 CNNR on (TRAIN-ALL) & .7186 & .7828 \\ [1ex] 
 \hline
\end{tabular}
\end{center}
\caption{results}
\end{table}

Insert table 5 from paper??

\subsection{Speech recognition}

\subsubsection{Convolutional Neural Networks for Speech Recognition}
In this paper Abdel-Hamid et al. describe how to apply CNNs to speech recognition in a novel way, such that the CNN?s structure directly accommodates some types of speech variability. They show a performance improvement relative to standard DNNs with similar numbers of weight parameters using this approach (about 6-10) relative error reduction), in contrast to the more equivocal results of convolving along the time axis, as earlier applications of CNNs to speech had attempted. Their hybrid CNN-HMM approach delegates temporal variability to the HMM, while convolving along the frequency axis creates a degree of invariance to small frequency shifts, which normally occur in actual speech signals due to speaker differences.they porpose a new, limited weight sharing scheme that can handle speech features in a better way than the full weight sharing that is standard in previous CNN architectures such as those used in image processing. Limited weight sharing leads to a much smaller number of units in the pooling layer, resulting in a smaller model size and lower computational complexity than the full weight sharing scheme.
An improved performance is observed on two ASR tasks: TIMIT phone recognition and a large-vocabulary voice search task, across a variety of CNN parameter and design settings. They determine that the use of energy information is very beneficial for the CNN in terms of recognition accuracy. Further, the ASR performance was found to be sensitive to the pooling size, but insensitive to the overlap between pooling units, a discovery that will lead to better efficiency in storage and computation. Finally, pretraining of CNNs based on convolutional RBMs was found to yield better performance in the large-vocabulary voice search experiment, but not in the phone recognition experiment. The results obtained for this paper can be seen in Table 9.

\begin{table*}[ht]
\begin{center}
 \begin{tabular}{||l l c c c c||} 
 \hline
 ID & Network structure & Average PER & min-max PER & params & ops \\ [0.5ex] 
 \hline\hline
 1 & DNN {2000 + 2x1000} & 22.02 & 21.86-22.11 & 6.9M & 6.9M \\ 
 \hline
 2 & DNN {2000 + 4x1000} & 21.87 & 21.68-21.98 & 8.9M & 8.9M \\
 \hline
 3 & CNN {LWS + 2x1000} & 20.17 & 19.92-20.41 & 5.4M & 10.7M \\
 \hline
 4 & CNN {FWS + 2x1000} & 20.31 & 20.16-20.58 & 8.5M & 13.6M \\
 \hline
 5 & CNN {FWS + FWS + 2x1000} & 20.23 & 20.11-20.29 & 4.5M & 11.7M \\
 \hline
 6 & CNN {FWS + LWS + 2x1000} & 20.36 & 19.91-20.61 & 4.1M & 7.5M \\ [1ex] 
 \hline
\end{tabular}
\end{center}
\caption{results}
\end{table*}

\subsubsection{Analysis of CNN-based Speech Recognition System using Raw Speech as Input}
In this paper Palaz et al. analyze CNN to understand the speech information that is modeled between the first two convolution layers. To that end, they present a method to compute the mean frequency responses of the filters in the first convolution layer that match to the specific inputs representing vowels. Studies on TIMIT task indicate that the mean frequency response tends to model the envelope of the sub-segmental (2-4 ms) speech signal. Then, they present a study to evaluate the susceptibility of the CNN-based system to mismatched conditions. This is an open problem in systems trained in a data-driven manner. They investigate this aspect on two tasks, namely, TIMIT phoneme recognition task and Aurora2 connected word recognition task. Our studies show that the performance of the CNN-based system degrades with the decrease in signal-to-noise ratio (SNR) like in a standard spectral feature based system. However, when compared to the spectral feature based system, the CNN-based system using raw speech signal as input yields better performance. The results obtained for this paper can be seen in Table 10.

\begin{table}[ht]
\begin{center}
 \begin{tabular}{||l l c c c c||} 
 \hline
 SNR [dB] & ANN & ANN & CNN & CNN \\ [0.5ex] 
 \hline\hline
 Training & clean & multi & clean & multi \\ 
 \hline
 2 & 52.5 & 54.3 & 65.5 & 66.8 \\
 \hline
 3 & 46.7 & 50.8 & 59.7 & 64.8 \\
 \hline
 4 & 40.3 & 46.6 & 50.5 & 60.8 \\
 \hline
 5 & 32.7 & 41.1 & 39.1 & 53.5 \\
 \hline
 5 & 26.1 & 34.2 & 27.8 & 42.8 \\
 \hline
 5 & 21.2 & 26.4 & 18.3 & 30.8 \\
 \hline
 6 & 17.4 & 20.2 & 9.9 & 21.4 \\ [1ex] 
 \hline
\end{tabular}
\end{center}
\caption{results}
\end{table}

\subsubsection{End-to-End Deep Neural Network for Automatic Speech Recognition}
In this paper Song et al. implement an end-to-end deep learning system that utilizes mel-filter bank features to directly output to spoken phonemes without the need of a traditional Hidden Markov Model for decoding. The system comprises of two variants of neural networks for phoneme recognition. In particular, a CNN is used for frame level classification and recurrent architecture with Connectionist Temporal Classification loss for decoding the frames into a sequence of phonemes. CNNs are exceptionally good at capturing high level features in spatial domain and have demonstrated unparalleled success in computer vision related tasks. One natural advantage of using CNN is that it?s invariant against translations of the variations in frequencies, which are common observed across speaker with different pitch due to their age or gender.
For each frame, the actual input is generated to the CNN by taking a window of frames surrounding it. Each input instance is a small one-channel image patch. The CNN architecture closely resembles many of architectures seen in recent years of research. (It consists of 4 convolutional layers where the first two layers have max pooling. After the convolutions, it's followed by two densely connected layer and finally a softmax layer. ReLU is used for all activation functions). One aspect where they differ is that instead of using the typical square convolution kernel, they use rectangular kernels since given a short window of frames, much of the information is stored across the frequency domain rather than the time domain.

\subsubsection{Applying Convolutional Neural Networks Concepts to Hybrid NN-HMM Model for Speech Recognition}
In this paper, Abdel-Hamid et al. propose to apply CNN to speech recognition within the framework of hybrid NN-HMM model. They propose to use local filtering and max-pooling in frequency domain to normalize speaker variance to achieve higher multi-speaker speech recognition performance. In their method, a pair of local filtering layer and max-pooling layer is added at the lowest end of neural network (NN) to normalize spectral variations of speech signals. Wit the use of the CNN they wish to normalize speech spectral features to achieve speaker invariance and enforce locality of features. The novelty of this paper is to apply the CNN concepts in the frequency domain to exploit CNN invariance to small shifts along the frequency axis through the use of local filtering and max-pooling. In this way, some acoustic variations can be effectively normalized and the resultant feature representation may be immune to speaker variations, colored background and channel noises. The results obtained for this paper can be seen in Table 11.

\begin{table}[ht]
\begin{center}
 \begin{tabular}{||l c||} 
 \hline
 Method & PER \\ [0.5ex] 
 \hline\hline
 NN with 3 hidden layers of 1000 nodes & 22.95 \\ 
 \hline
 CNN with no pre-training (their work) & 20.07 \\
 \hline
 NN with DBN pre-training & 20.70 \\
 \hline
 NN with DBN pre-training and mcRBM features extraction & 20.50 \\ [1ex] 
 \hline
\end{tabular}
\end{center}
\caption{results}
\end{table}

\section{Journals}
\subsection{Classifying Relations by Ranking with Convolutional Neural Networks (P15-1061)}
In this work, Dong et al. propose a new convolutional neural network (CNN), named Classification by Ranking CNN (CR-CNN), to tackle the relation classification task. The proposed network learns a distributed vector representation for each relation class. Given an input text segment, the network uses a convolutional layer to produce a distributed vector representation of the text and compares it to the class representations in order to produce a score for each class. They propose a new pairwise ranking loss function that makes it easy to reduce the impact of artificial classes. Using CRCNN, and without the need for any costly handcrafted feature, they outperform the state-of-the-art for the SemEval-2010 Task 8 dataset. Their experimental results are evidence that: 
1) CR-CNN is more effective than CNN followed by a softmax classifier. 
2) Omitting the representation of the artificial class Other improves both precision and recall. 
3) Using only word embeddings as input features is enough to achieve state-of-the-art results if only the text between the two target nominals is considered. 
The results obtained for this paper can be seen in Table 12.

\begin{table*}[ht]
\begin{center}
 \begin{tabular}{||l p{4in} c||} 
 \hline
 Classifier & Feature Set & F1 \\ [0.5ex] 
 \hline\hline
 SVM (Rink and Harabagiu, 2010) & POS, prefixes, morphological, WordNet, dependency parse, Levin classes, ProBank, FrameNet, NomLex-Plus,Google n-gram, paraphrases, TextRunner & 82.2 \\ 
 \hline
 RNN (Socher et al., 2012) & word embeddings & 74.8 \\
 \hline
 RNN (Socher et al., 2012) & word embeddings, POS, NER, WordNet & 77.6 \\
 \hline
 MVRNN (Socher et al., 2012) & word embeddings & 79.1 \\
 \hline
 MVRNN (Socher et al., 2012) & word embeddings, POS, NER, WordNet & 82.4 \\
 \hline
 CNN+Softmax (Zeng et al., 2014) & word embeddings & 69.7 \\
 \hline
 CNN+Softmax (Zeng et al., 2014) & word embeddings, word position embeddings, word pair, words around word pair, WordNet & 82.7 \\
 \hline
 FCM (Yu et al., 2014) & word embeddings & 80.6 \\
 \hline
 FCM (Yu et al., 2014) & word embeddings, dependency parse, NER & 83.0 \\
 \hline
 CR-CNN & word embeddings & 32.8 \\
 \hline
 CR-CNN & word embeddings, word position embeddings & 84.1 \\ [1ex] 
 \hline
\end{tabular}
\end{center}
\caption{Results}
\end{table*}

\subsection{A Convolutional Architecture for Word Sequence Prediction. P15-1151 (genCNN, difficult to understand)1/2 I have to add more info, but I'm having troubles to understand it.}
In this paper, et al. propose a novel convolutional architecture, named genCNN, as a model that can efficiently combine local and long range structures of language for the purpose of modeling conditional probabilities. genCNN can be directly used in generating a word sequence (i.e., text generation) or evaluating the likelihood of word sequences (i.e., language modeling). They also show the empirical superiority of genCNN on both tasks over traditional n-grams and its RNN or FFN counterparts. The results obtained for this paper can be seen in Table 13.

\begin{table}[ht]
\begin{center}
 \begin{tabular}{||c c c c||} 
 \hline
 Models & MT06 & MT08 & Average \\ [0.5ex] 
 \hline\hline
 Baseline & 38.63 & 31.11 & 34.87 \\ 
 \hline
 RNN rerank & 39.03 & 31.50 & 35.26 \\
 \hline
 LSTM rerank & 39.20 & 31.90 & 35.55 \\
 \hline
 FFN-LM rerank & 38.93 & 31.41 & 35.14 \\
 \hline
 genCNN rerank & 39.90 & 32.50 & 36.20 \\
 \hline
 Base+FFN-LM & 39.08 & 31.60 & 35.34 \\
 \hline
 genCNN rerank & 40.4 & 32.85 & 36.63 \\ [1ex] 
 \hline
\end{tabular}
\end{center}
\caption{results}
\end{table}
\subsection{A Convolutional Neural Network for Modelling Sentences (P14-1062)}
In this paper Kalchbrenner et al. use the Dynamic Convolutional Neural Network (DCNN) for the semantic modeling of sentences. The network handles input sentences of varying length and induces a feature graph over the sentence that is capable of explicitly capturing short and long-range relations. Multiple layers of convolutional and dynamic pooling operations induce a structured feature graph over the input sentence. Small filters at higher layers can capture syntactic or semantic relations between non-continuous phrases that are far apart in the input sentence. The feature graph induces a hierarchical structure somewhat akin to that in a syntactic parse tree. The structure is not tied to purely syntactic relations and is internal to the neural network.
They experiment with the network in four settings. The first two experiments involve predicting the sentiment of movie reviews. The network outperforms other approaches in both the binary and the multi-class experiments. The third experiment involves the categorization of questions in six question types. The fourth experiment involves predicting the sentiment of Twitter posts using distant supervision. The network is trained on 1.6 million tweets labelled automatically according to the emoticon that occurs in them.

\subsection{Sequential Short-Text Classification with Recurrent and Convolutional Neural Networks(N16-1062)}
In this work, Lee et al. present a model based on recurrent neural networks and convolutional neural networks that incorporates the preceding short texts. Inspired by the performance of ANN-based systems for non-sequential short-text classification, they introduce a model based on recurrent neural networks (RNNs) and CNNs for sequential short-text classification, and evaluate it on the dialog act classification task. A dialog act characterizes an utterance in a dialog based on a combination of pragmatic, semantic, and syntactic criteria. Its accurate detection is useful for a range of applications, from speech recognition to automatic summarization. Their model comprises two parts. The first part generates a vector representation for each short text using either the RNN or CNN architecture. The second part classifies the current short text based on the vector representations of the current as well as a few preceding short texts. The results obtained for this paper can be seen in Table 14.

\begin{table}[ht]
\begin{center}
 \begin{tabular}{||l c c c||} 
 \hline
 Models & DSTC 4 & MRDA & SwDA \\ [0.5ex] 
 \hline\hline
 CNN & 65.5 & 84.6 & 73.1 \\ 
 \hline
 LSTM & 66.2 & 84.3 & 69.6 \\
 \hline
 Majority class & 25.8 & 59.1 & 33.7 \\
 \hline
 SVM & 57.0 & - & - \\
 \hline
 Graphical model & - & 81.3 & - \\
 \hline
 Naive Bayes & - & 82.0 & - \\
 \hline
 HMM & - & - & 71.0 \\
 \hline
 Memory-based Learning & - & - & 72.3 \\
 \hline
 Interlabeler agreement & - & - & 84.0 \\ [1ex] 
 \hline
\end{tabular}
\end{center}
\caption{results}
\end{table}

\section{Conclusion}
This paper presents state-of-the-art deep learning tools for Natural Language Processing. 
The main contributions of this work are??An overview of CNN and its different subtypes. A get together of all the problems that have been solved using state-of-the-art CNN technologies. A general view of how CNN have been applied to different NLP problems, with results included. 

After the advances made in Computer Vision using deep learning tools, NLP has adapted some of these techniques to make major breakthroughs. However, the results, for now, are only promising. There is evidence that deep learning tools provide good solutions, but they haven't provided such a big leap as the one in Computer Vision.

One of the main problems is that CNN started being used because of the great success in CV. Due to this there's a lack of a common goal. This uncertainty of what to do causes the results to be good but not as good as expected. One of the reasons could be because CNN are thought to be applied to images and not to words.
However, the results and all the ... are encouraging and are an improvement over the previous state-of-the-art techniques.

\section{Future Work}
There's a need to define common goals and set a better use of CNN. 
Convolutional Neural Networks are designed to be used on images. Missing component (2D-3D)

Speech recognition seems the area with the best results (maybe because it's one of the areas that concerns a bigger number of people). Try to see the model they have used and adapt it to the problem the author is trying to solve.


%

\balance

\section*{Acknowledgment}

The authors would like to thank...

\ifCLASSOPTIONcaptionsoff
  \newpage
\fi




\begin{thebibliography}{28}

\bibitem{IEEEhowto:zong}
J. Zhang and C. Zong, \emph{Deep Neural Networks in Machine Translation: An Overview}, IEEE Intell. Syst., 2015.

\bibitem{IEEEhowto:zhu}
C. Zhu, X. Qiu, X. Chen, and X. Huang, \emph{A Re-ranking Model for Dependency Parser with Recursive Convolutional Neural Network}, Proc. 53rd Annu. Meet. Assoc. Comput. Linguist. 7th Int. Jt. Conf. Nat. Lang. Process. Volume 1 Long Pap., pp. 1159?1168, 2015.

\bibitem{IEEEhowto:wangp}
P. Wang, J. Xu, B. Xu, C. Liu, H. Zhang, F. Wang, and H. Hao, \emph{Semantic Clustering and Convolutional Neural Network for Short Text Categorization}, Proc. ACL 2015, pp. 352?357, 2015.
  
\bibitem{IEEEhowto:francis}
M. Francis-Landau, G. Durrett, and D. Klein, \emph{Capturing Semantic Similarity for Entity Linking with Convolutional Neural Networks}, pp. 1256?1261, 2016. 

\bibitem{IEEEhowto:zhang}
R. Zhang, H. Lee, and D. Radev, \emph{Dependency Sensitive Convolutional Neural Networks for Modeling Sentences and Documents}, Naacl-Hlt-2016, pp. 1512?1521, 2016.
  
\bibitem{IEEEhowto:chen}
Y. Chen, L. Xu, K. Liu, D. Zeng, and J. Zhao, \emph{Event Extraction via Dynamic Multi-Pooling Convolutional Neural Networks}, Proc. ACL 2015, pp. 167?176, 2015. 

\bibitem{IEEEhowto:nguyen}
T. H. Nguyen and R. Grishman, \emph{Event Detection and Domain Adaptation with Convolutional Neural Networks}, Proc. 53rd Annu. Meet. Assoc. Comput. Linguist. 7th Int. Jt. Conf. Nat. Lang. Process. (Volume 2 Short Pap., pp. 365?371, 2015.

\bibitem{IEEEhowto:vu}
N. T. Vu, H. Adel, P. Gupta, and H. Schutze, \em{Combining Recurrent and Convolutional Neural Networks for Relation Classification}, pp. 534?539, 2016.

\bibitem{IEEEhowto:naacl}
A. Naacl, \emph{Comparing Convolutional Neural Networks to Traditional Models for Slot Filling} pp. 1?9, 2016.
  
\bibitem{IEEEhowto:denil}
M. Denil, A. Demiraj, N. Kalchbrenner, P. Blunsom, and N. de Freitas, \emph{Modelling, Visualising and Summarising Documents with a Single Convolutional Neural Network}, arXiv Prepr. arXiv1406.3830, pp. 1?10, 2014.  

\bibitem{IEEEhowto:hu}
B. Hu, Z. Tu, Z. Lu, and Q. Chen, \emph{Context-Dependent Translation Selection Using Convolutional Neural Network}, Proc. 53rd Annu. Meet. Assoc. Comput. Linguist. 7th Int. Jt. Conf. Nat. Lang. Process. (Volume 1 Long Pap., no. Section 3, pp. 536?541, 2015.

\bibitem{IEEEhowto:meng}
F. Meng, Z. Lu, M. Wang, H. Li, W. Jiang, and Q. Liu, \emph{Encoding Source Language with Convolutional Neural Network for Machine Translation}, Proc. 53rd Annu. Meet. Assoc. Comput. Linguist. 7th Int. Jt. Conf. Nat. Lang. Process. (Volume 1 Long Pap., pp. 20?30, 2015.

\bibitem{IEEEhowto:dong}
L. Dong, F. Wei, M. Zhou, and K. Xu, \emph{Question Answering over Freebase with Multi-Column Convolutional Neural Networks}, Proc. ACL 2015, pp. 260?269, 2015.

\bibitem{IEEEhowto:severyn}
A. Severyn and A. Moschitti, \emph{Modeling Relational Information in Question-Answer Pairs with Convolutional Neural Networks}, Arxiv, 2016.

\bibitem{IEEEhowto:palaz}
D. Palaz, M. Magimai-Doss, and R. Collobert, \emph{Analysis of CNN-based speech recognition system using raw speech as input}, Proc. Annu. Conf. Int. Speech Commun. Assoc. INTERSPEECH, vol. 2015-Janua, pp. 11?15, 2015.

\bibitem{IEEEhowto:song}
W. Song and J. Cai, \emph{End-to-End Deep Neural Network for Automatic Speech Recognition}, pp. 1?8, 2015.

\bibitem{IEEEhowto:abdel}
O. Abdel-hamid, H. Jiang, and G. Penn \emph{Applying  convolutional neural networks concepts to hybrid NN-HMM model for speech recognition}, Department of Computer Science and Engineering, York University, Toronto, Canada Department of Computer Science, University of Toronto, Toronto, Canada, Acoust. Speech Signal Process. (ICASSP), 2012 IEEE Int. Conf., pp. 4277?4280, 2012.

\bibitem{IEEEhowto:santos}
C. N. dos Santos, B. Xiang, and B. Zhou, \emph{Classifying Relations by Ranking with Convolutional Neural Networks}, Acl-2015, no. 3, pp. 626?634, 2015.

\bibitem{IEEEhowto:wangm}
M. Wang, Z. Lu, H. Li, W. Jiang, and Q. Liu, \emph{A Convolutional Architecture for Word Sequence Prediction}, Acl-2015, pp. 1567?1576, 2015.

\bibitem{IEEEhowto:kalch}
N. Kalchbrenner, E. Grefenstette, and P. Blunsom, \emph{A Convolutional Neural Network for Modelling Sentences}, Proc. 52nd Annu. Meet. Assoc. Comput. Linguist. (ACL 2014), pp. 655?665, 2014.

\bibitem{IEEEhowto:lee}
J. Y. Lee and F. Dernoncourt, \emph{Sequential Short-Text Classification with Recurrent and Convolutional Neural Networks}, Naacl, pp. 515?520, 2016.





\bibitem{IEEEhowto:bitvai}
Z. Bitvai and T. Cohn, ?Non-Linear Text Regression with a Deep Convolutional Neural Network,? Proc. ACL 2015, pp. 180?185, 2015.

\bibitem{IEEEhowto:cao}
Z. Cao, F. Wei, S. Li, W. Li, M. Zhou, and H. Wang, ?Learning Summary Prior Representation for Extractive Summarization,? Proc. ACL 2015, pp. 829?833, 2015.

\bibitem{IEEEhowto:golik}
P. Golik, Z. Tuske, R. Schuler, and H. Ney, Convolutional neural networks for acoustic modeling of raw time signal in LVCSR, Proc. Annu. Conf. Int. Speech Commun. Assoc. INTERSPEECH, vol. 2015-Janua, pp. 26-30, 2015.

\bibitem{IEEEhowto:king}
B. King, R. Jha, T. Johnson, and V. Sundararajan, ?Experiments in Automatic Text Summarization Using Deep Neural Networks, March. , 2011.

\bibitem{IEEEhowto:ma}
M. Ma, L. Huang, B. Xiang, and B. Zhou, ?Dependency-based Convolutional Neural Networks for Sentence Embedding,? Acl-2015, no. 1995, pp. 174?179, 2015.

\bibitem{IEEEhowto:sainath}
T. N. Sainath, A. R. Mohamed, B. Kingsbury, and B. Ramabhadran, ?Deep convolutional neural networks for LVCSR,? ICASSP, IEEE Int. Conf. Acoust. Speech Signal Process. - Proc., pp. 8614?8618, 2013.

\bibitem{IEEEhowto:thomas}
S. Thomas, S. Ganapathy, G. Saon, and H. Soltau, ?Analyzing convolutional neural networks for speech activity detection in mismatched acoustic conditions,? ICASSP, IEEE Int. Conf. Acoust. Speech Signal Process. - Proc., pp. 2519?2523, 2014.



\bibitem{IEEEhowto:kopka}
H.~Kopka and P.~W. Daly, \emph{A Guide to kfnng}, 3rd~ed.\hskip 1em plus
  0.5em minus 0.4em\relax Harlow, England: Addison-Wesley, 1999.

\end{thebibliography}
%

%

\begin{IEEEbiography}[{\includegraphics[width=1in,height=1.25in,clip,keepaspectratio]{picture}}]{John Doe}
\blindtext
\end{IEEEbiography}




\end{document}